\documentclass{llncs}
\pdfoutput=1
\usepackage[T1]{fontenc}
\usepackage{makeidx}

\usepackage{makeidx}
\usepackage{graphicx}
\usepackage{amssymb}
\usepackage{amsmath}
\usepackage{amscd}

\newcommand{\commentout}[1]{}

\newcommand{\introskip}{\vspace{3ex}}

\newcommand{\E}{\mathbb{E}}                    

\newcommand{\abs}[1]{\mathop{\left\lvert #1 \right\rvert}} 
\newcommand{\args}[1]{\mathop{\left( #1 \right)}} 

\newcommand{\norm}[1]{\mathop{\left\lVert #1 \right\rVert}}
\newcommand{\cbrace}[1]{\mathop{\left\{ #1 \right\}}}
\newcommand{\bracket}[1]{\mathop{\left[ #1 \right]}}

\newcommand{\argsS}[2]{\mathop{\left( #1 \right)#2}}

\newcommand{\T}{\mathop{\mathsf{T}}}           

\renewcommand{\S}[1]{{\mathcal{#1}}}           
\def\vec#1{\mathchoice{\mbox{\boldmath$\displaystyle#1$}}
{\mbox{\boldmath$\textstyle#1$}}
{\mbox{\boldmath$\scriptstyle#1$}}
{\mbox{\boldmath$\scriptscriptstyle#1$}}}

{%
\end{array}\right.}
\newcommand{\atab}[1]{\hspace*{#1em}}

\newcounter{algorithm_counter}
\newenvironment{algorithm}[1]{
\refstepcounter{algorithm_counter}
\setlength{\parindent}{0\parindent}
\vspace{2ex}
\begin{minipage}{\textwidth}
\rule{\textwidth}{5\arrayrulewidth}\\
\begin{footnotesize}
{\bf Algorithm \arabic{algorithm_counter}:} #1 \\
\rule[+1.5ex]{\textwidth}{\arrayrulewidth}

\vspace{-1.5ex}

}%
{
\\[-1.5ex]
\rule{\textwidth}{\arrayrulewidth}
\end{footnotesize}
\end{minipage}
\setlength{\parindent}{\parindent}
}

{
\\[-1.5ex]
\rule{\textwidth}{\arrayrulewidth}
\end{footnotesize}
\end{minipage}
\setlength{\parindent}{\parindent}
}

\begin{document}
\title{Accelerating Competitive Learning Graph Quantization}
\titlerunning{}

\author{Brijnesh J.~Jain$^1$  \and Klaus Obermayer$^1$} 
\authorrunning{Brijnesh J.~Jain and Klaus Obermayer}  
\institute{$^1$Berlin Institute of Technology, Germany\\
\email{\{jbj|oby\}@cs.tu-berlin.de}}

\maketitle 

\begin{abstract}  
Vector quantization(VQ) is a lossy data compression technique from signal processing for which simple competitive learning is one standard method to quantize patterns from the input space. Extending competitive learning VQ to the domain of graphs results in competitive learning for quantizing input graphs. In this contribution, we propose an accelerated version of competitive learning graph quantization (GQ) without  trading computational time against solution quality. For this, we lift graphs locally to vectors in order to avoid unnecessary calculations of intractable graph distances. In doing so, the accelerated version of competitive learning GQ gradually turns locally into a competitive learning VQ with increasing number of iterations. Empirical results show a significant speedup by maintaining a comparable solution quality.
\end{abstract}

\section{Introduction}

Vector quantization is a classical technique from signal processing suitable for lossy data compression, density estimation, and prototype-based clustering \cite{Gersho92}. The aim of optimal vector quantizer design is to find a codebook of a given size $k$ such that an expected distortion with respect to some (differentiable) distortion measure is minimized. Competitive learning VQ is one standard method for optimal vector quantizer design \cite{Duda00,Hertz91,Theodoridis09}. 

Graph quantization as formalized in \cite{Jain09b} is a generalization of vector quantization to the quantization of combinatorial structures that can be represented by attributed graphs. The generalization from vectors to graphs opens applications to diverse domains like proteomics, chemoinformatics, and computer vision, where the patterns to be quantized are  more naturally represented by trees, lattices, and graphs.

Designing optimal graph quantizers has been pursued as central clustering. Examples include competitive learning algorithms in the domain of graphs \cite{Gold96b,Guenter02,Jain04,Jain09a} and k-means as well as k-medoids algorithms  \cite{Ferrer07,Ferrer09,Jain04,Jain08,Jain09b,Jain09c,Schenker03,Schenker05}. A key problem of all these algorithms is that they are slow in practice for large datasets of graphs, because calculating the underlying graph distance (distortion) measure is a graph matching problem of exponential complexity. Given a training set of $N$ graphs, competitive learning, k-means, and k-median calculate (or approximate) at least $kN$ intractable graph distances during each cycle through the training set. 

In this contribution, we propose an accelerated version of competitive learning GQ. We assume that the underlying distortion measure is a geometric graph metric, which arises in various different guises as a common choice of proximity measure in a number of applications \cite{Almohamad93,Caetano07,Cour06,Gold96a,Umeyama1988,Wyk02}. The proposed accelerated version of competitive learning GQ avoids unnecessary graph distance calculations by exploiting metric properties and by lifting the graphs isometrically to an Euclidean vector space whenever possible. Lifting graphs to vectors reduces competitive learning GQ to competitive learning VQ locally. If lifting becomes unfeasible, we switch back to competitive learning GQ. By switching back and forth between competitive learning GQ and competitive learning VQ locally, the accelerated version of competitive learning GQ  reduces gradually to the efficient competitive learning algorithm for VQ.

 The proposed accelerated version of competitive learning GQ has the following properties: First, it can be applied to finite combinatorial structures other than graphs like, for example, point patterns, sequences, trees, and hypergraphs. For the sake of concreteness, we restrict our attention exclusively to the domain of graphs. Second, any initialization method that can be used for competitive learning GQ can also be used for its accelerated version. Third, hierarchical methods for GQ benefit from the proposed accelerated scheme. Fourth, competitive learning GQ and its accelerated version perform comparable with respect to solution quality. Different solutions are caused by approximation errors of the graph matching algorithm and by multiple local minima but are not caused by the mechanisms to accelerate competitive learning.
 
The paper is organizes as follows. Section 2 briefly describes competitive learning GQ. Section 3 proposes an accelerated version of competitive learning GQ. Experimental results are presented and discussed in Section 4. Finally, Section 5 concludes with a summary of the main results and future work.

\section{Competitive Learning Graph Quantization}

This section briefly introduces competitive learning as a design technique for graph quantization. For details on graph quantization and consistency results of competitive learning, we refer to \cite{Jain09b}.

\subsection{Metric Graph Spaces}

Let  $\E$ be a $r$-dimensional Euclidean vector space.  An (\emph{attributed}) \emph{graph} is a triple $X = (V,E,\alpha)$ consisting of a finite nonempty set $V$ of \emph{vertices}, a set $E \subseteq V \times V$ of \emph{edges}, and an \emph{attribute function} $\alpha:V\times V \rightarrow \E$, such that $\alpha(i,j) = \vec{0}$ for each pair of distinct vertices $i, j$ with $(i,j)\notin E$. 

For simplifying the mathematical treatment, we assume that all graphs are of order $n$, where $n$ is chosen to be sufficiently large. Graphs of order less than $n$, say $m < n$, can be extended to order $n$ by including isolated vertices with attribute zero. For practical issues, it is important to note that limiting the maximum order to some arbitrarily large number $n$ and extending smaller graphs to graphs of order $n$ are purely technical assumptions to simplify mathematics. For machine learning problems, these limitations should have no practical impact, because neither the bound $n$ needs to be specified explicitly nor an extension of all graphs to an identical order needs to be performed. When applying the theory, all we actually require is that the graphs are finite. 

A graph $X$ is completely specified by its \emph{matrix representation} $\vec{X} = (\vec{x}_{ij})$ with elements $\vec{x}_{ij} = \alpha(i,j)$ for all $1 \leq i,j\leq n$. By  concatenating the columns of $\vec{X}$, we obtain a \emph{vector representation} $\vec{x}$ of $X$. 

Let $\S{X} = \E^{n \times n}$ be the Euclidean space of all ($n \times n$)-matrices and let $\S{T}$ denote a subgroup of the set $\S{P}^n$ of all 
$(n\times n)$-permutation matrices. Two matrices $\vec{X}\in\S{X}$ and $\vec{X'}\in\S{X}$ are said to be equivalent, if there is a permutation matrix $P \in \S{T}$ such that $\vec{P}^{\T}\vec{X}\vec{P} = \vec{X'}$. The quotient set 
\[
\S{X_T} = \S{X}/\S{T} = \cbrace{[\vec{X}] \,:\, \vec{X} \in \S{X}}
\] 
is a \emph{graph space} of all \emph{abstract graphs} $[\vec{X}]$ over the \emph{representation space} $\S{X}$ induced by the transformation group $\S{T}$. Note that the graph space $\S{X_T}$ is a Riemannian orbifold. The notion of orbifold is fundamental to extend analytical and geometrical properties of Euclidean spaces to graph spaces. For details, we refer to \cite{Jain09b}.

In the remainder of this contribution, we identify $\S{X}$ with $\E^N$ ($N = n^2$) and consider vector- rather than matrix representations of abstract graphs. By abuse of notation, we sometimes identify $X$ with $[\vec{x}]$ and write $\vec{x} \in X$ instead of $\vec{x} \in \bracket{\vec{x}}$.  We say, $X$ \emph{lifts} to $\vec{x}$ if we represent graph $X$ by  vector $\vec{x} \in X$.

Finally, we equip our graph space with a metric. Let $\norm{\cdot}$ be a Euclidean norm on $\S{X}$. Then the distance function
\[
d(X, Y) = \min \cbrace{\norm{\vec{x} - \vec{y}} \,:\, \vec{x} \in X, \vec{y} \in Y},
\] 
is a metric. It is well known that calculating the graph distance metric $d(X,Y)$ is a NP-complete graph matching problem \cite{Gold96a}.

Since $\S{T}$ is a group, we have 
\[
d_{\vec{y}}(X) = \min \cbrace{\norm{\vec{x} - \vec{y}} \,:\, \vec{x} \in X} = d(X, Y), 
\]
where $\vec{y}$ is an arbitrary vector representation of $Y$. By symmetry, we similarly have $d_{\vec{x}}(Y) = d(Y,X)$.  Hence, the graph distance $d(X, Y)$ can be determined by fixing an arbitrary vector representation $\vec{y}\in Y$ and then finding a vector representation from $X$ that minimizes $\norm{\vec{x} - \vec{y}}$ over all vector representations $\vec{x}$ from $X$.

A pair $(\vec{x}, \vec{y}) \in X \times Y$ of vector representations is called \emph{optimal alignment} if $\norm{\vec{x} - \vec{y}} = d(X, Y)$. Thus, we have $d(X, Y) \leq \norm{\vec{x} - \vec{y}}$ for all vector representations $\vec{x} \in X$ and $\vec{y} \in Y$, where equality holds if and only if $\vec{x}$ and $\vec{y}$ are optimally aligned.

\subsection{Graph Quantization}

Let $\args{\S{X_T}, d}$ be a metric graph space over the Euclidean space $(\S{X}, \norm{\cdot})$. A \emph{graph quantizer} of size $k$ is a mapping of the form
\[
Q : \S{X_T} \rightarrow \S{Y_T},
\]
where $\S{C} = \cbrace{Y_1, \ldots, Y_k} \subseteq \S{X_T}$ is a \emph{codebook}. The elements $Y_j\in\S{Y_T}$ are the \emph{code graphs}. 

Adaptive graph quantization design aims at finding a codebook $\S{C} = \cbrace{Y_1, \ldots, Y_k} \subseteq \S{X_T}$ such that the \emph{expected distortion}
\begin{align*}
D\args{\S{C}} &= \sum_{j=1}^{k} \int_{\S{X_T}} \min_{1 \leq j \leq k} d\argsS{X, Y_j}{^2} f(X) dX
\end{align*}
is minimized, where $\S{C} \in \S{X}_{\S{T}}^k$ and $f = f_{\S{X_T}}$ is a probability density defined on some measurable space $\args{\S{X_T}, \Sigma_{\S{X_T}}}$.

A statistically consistent method to minimize the expected distortion using empirical data is competitive learning as outlined in Algorithm \ref{alg:CLGQ}.
\begin{figure}[tbp]
\centering
\begin{algorithm}{Competitive Learning GQ}\label{alg:CLGQ}
01 \atab{1} choose an initial codebook $\S{C} = \cbrace{Y_{1}, \ldots, Y_{k}}\subseteq \S{X_T}$\\
02 \atab{1} choose arbitrary vector representations $\vec{y}_{1}\in Y_1, \ldots, \vec{y}_{k} \in Y_k$\\
03 \atab{1} \textbf{repeat} \\
04 \atab{3} randomly select an input graph $X \in \S{X_T}$ \\
05 \atab{3} let $Y_X  = \arg\min_{Y \in \S{C}} d\!\args{X,Y}{^{\!2}}$\\
06 \atab{3} choose $\vec{x} \in X$ optimally aligned with $\vec{y} \in Y_{X}$\\
07 \atab{3} determine learning rate $\eta_i > 0$\\
08 \atab{3} update $\vec{y} = \vec{y} + \eta \args{\vec{x}-\vec{y}}$\\
09 \atab{1} \textbf{until} some termination criterion is satisfied
\end{algorithm}
\end{figure}

\section{Accelerating Competitive Learning GQ}

This section proposes an acceleration of competitive learning GQ that is based on a fixed training set $\S{S}= \cbrace{X_{1}, \ldots, X_{N}}$ consisting of $N$ independent graphs $X_i$ drawn from $\S{X}_\Gamma$. At each cycle through the training set, competitive learning GQ as described in Algorithm \ref{alg:CLGQ} calculates $kN$ graph distances, each of which is NP-hard. These distance calculations predominate the computational cost of Algorithm \ref{alg:CLGQ}. Accelerating competitive learning GQ therefore aims at reducing the number of graph distance calculations. 

\introskip

Accelerating competitive learning GQ is based on the key idea that a small change of a vector representation $\vec{y} \in Y$ of a code graph $Y\in \S{C}$ does not change  already optimally aligned vector representations $\vec{x}\in X$ of training graphs $X\in \S{S}$ quantized by (closest to) $Y$. When keeping track of the most recent optimal alignment, calculating graph distances reduces to calculating Euclidean distances of vector representations, in particular in the final convergence phase. Thus, in order to avoid graph distance calculations, we lift graphs to their optimally aligned vector representations --- if possible --- and switch to Euclidean distance calculations.

To describe the accelerated version of Algorithm  \ref{alg:CLGQ}, we assume that $X \in \S{S}$ is a training graph and $Y, Y_X \in \S{C}$ are code graphs. By $Y_X = Q(X)$ we denote the \emph{encoding} of $X$. Since the graph distance $d$ is a metric, we have
\begin{align}\label{eq:u<l}
u(X) \leq l(X, Y) \;\Rightarrow\; d\!\args{X, Y_X} \leq d\!\args{X, Y},
\end{align}
where 
\begin{enumerate}
\item $u(X) \geq d\!\args{X, Y_X}$ denotes an upper bound of  $d\!\args{X, Y_X}$ and
\item $l(X, Y) \leq d\!\args{X, Y}$ denotes a lower bound of $d\!\args{X, Y}$.
\end{enumerate}
From Eqn.\ (\ref{eq:u<l}) follows that we can avoid calculating a graph distance $d\!\args{X, Y}$ if at least one of the following two conditions is satisfied
\begin{description}
\item[$(C_1)$] $Y = Y_X$
\item[$(C_2)$] $u(X) \leq l(X, Y)$.
\end{description}
A code graph $Y$ is a \emph{candidate encoding} for $X$ if both conditions $(C_1)$ and $(C_2)$ are violated. In this case, we apply the technique of \emph{delayed distance evaluation}. For this, we proceed as follows:
\begin{enumerate}
\item We first test whether the upper bound $u(X)$ is out-of-date. An upper bound is out-of-date if $u(X) \gneqq d\!\args{X, Y_X}$. Note that we can check the state of $u(X)$ without calculating the distance $d\!\args{X, Y_X}$. 
\item If $u(X)$ is out-of-date, we calculate the distance $d\!\args{X, Y_X}$ and set $u(X) = d\!\args{X, Y_X}$. Since improving the upper bound $u(X)$ might eliminate $Y$ as being a candidate encoding for $X$, we recheck condition $(C_2)$. 
\item If condition $(C_2)$ is still violated despite the updated upper bound $u(X)$, we have the following situation
\[
u(X) = d\!\args{X, Y_X} > l(X, Y).
\]
We update the lower bound by calculating $l(X,Y) = d\!(X, Y)$ and then re-examine condition $(C_2)$.
\item If condition $(C_2)$ remains violated, we have
\[
u(X) = d\!\args{X, Y_X} > d(X, Y) = l(X, Y). 
\]
This implies that $X$ is closer to code graph  $Y$ than to its current encoding $Y_X$. In this case $Y$ becomes the new encoding of $X$. 
\end{enumerate}
Otherwise, if at least one of both conditions $(C_1)$ and $(C_2)$ is satisfied or becomes satisfied during examination, $Y_X$ remains the encoding of $X$.

\introskip

\begin{figure}[htpb]
\centering
\begin{algorithm}{Accelerated Competitive Learning GQ}\label{alg:ACLGQ}
01 \atab{1} choose an initial codebook $\S{C} = \cbrace{Y_{1}, \ldots, Y_{k}}\subseteq \S{X_T}$\\
02 \atab{1} choose arbitrary vector representations $\vec{y}_{1}\in Y_1, \ldots, \vec{y}_{k} \in Y_k$\\
03 \atab{1} set $u(X) = \infty$ and declare $u(X)$ out-of-date for all $X\in \S{S}$\\
04 \atab{1} \textbf{repeat} \\
05 \atab{3} store $Y$ in $Y'$ for  all $Y\in \S{C}$\\
06 \atab{3} \textbf{repeat} \\
07 \atab{5} randomly select $X \in \S{S}$ \\
08 \atab{5} $Y_X$ = \textsc{classify}(X)\\
09 \atab{5} determine learning rate $\eta > 0$\\
10 \atab{5} set $\vec{y} = \vec{y} + \eta \args{\vec{x}_a-\vec{y}}$\\
11 \atab{3} \textsc{estimate\_bounds}()\\
12 \atab{1} \textbf{until} some termination criterion is satisfied
\end{algorithm}
\vspace{-0.2cm}
\begin{algorithm}{\textsc{classify}(X)}\label{alg:ACLGQ:classify}
01 \atab{1} \textbf{for} each $Y \in \S{C}$ \textbf{do}\\
02 \atab{3} \textbf{if} $Y$ is a candidate encoding for $X$\\
03 \atab{5} \textbf{if} $u\args{X}$ is out-of-date \\
04 \atab{7} \textsc{update\_bounds(X, Y)}\\
05 \atab{5} \textbf{if} $Y$ is still a candidate encoding for $X$\\
06 \atab{7} \textbf{if} $d\!\args{X, Y} < u\args{X}$\\
07 \atab{9} \textsc{update\_bounds(X, Y)}\\
08 \atab{9} set $Y_X = Y$\\
09 \atab{1} \textbf{return} $Y_X$
\end{algorithm}
\vspace{-0.5cm}
\begin{algorithm}{\textsc{update\_bounds(X, Y)}}\label{alg:ACLGQ:updateBounds}
01 \atab{1} set $u(X) = d\!\args{X, Y_X}$\\
02 \atab{1} set $l\args{X, Y_X} = d\!\args{X, Y_X}$\\
03 \atab{1} declare $u(X)$ as up-to-date\\
04 \atab{1} store $\vec{x}_a\in X$ with $d\!\args{X, Y_X} = \norm{\vec{x}_a -\vec{y}}$
\end{algorithm}
\vspace{-0.5cm}
\begin{algorithm}{\textsc{estimate\_bounds}()}\label{alg:ACLGQ:estimateBounds}
01 \atab{1} compute $\delta(Y) = d\!\args{Y, Y'}$  for all $Y\in \S{C}$\\
02 \atab{1} set $u(X) = \min\cbrace{u(X) + \delta\args{Y_X}, \norm{\vec{x}_a - \vec{y}}}$ for all $X\in \S{X_T}$\\
03 \atab{1} declare $u(X)$ as out-of-date for all $X\in \S{X_T}$\\
04 \atab{1} set $l(X,Y) = \max\cbrace{l(X,Y') - \delta\args{Y}, 0}$ for all $X\in \S{X_T}$ and for all $Y\in \S{C}$
\end{algorithm}
\end{figure}

Crucial for avoiding NP-hard graph distance calculations are good estimates of the lower and upper bounds $l(X, Y)$ and $u(X)$ after each cycle through the training set $\S{S}$. For this, we keep record of the most recent optimal alignment of an input graph $X$ and its current code graph $Y_X$. Suppose that $\vec{y}$ is a vector representation of $Y_X$ during a cycle through the training set. Each time we calculate a distance $d\!\args{X, Y_X}$, we obtain a vector representation $\vec{x} \in X$ such that $\args{\vec{x}, \vec{y}}$ is an optimal alignment of $X \times Y_X$. By $\vec{x}_a \in X$ we denote the vector representation of $X$ of an optimal alignment obtained by the most recent distance calculation $d\!\args{X,Y_X}$. Updating the bounds is then carried out as follows: After each cycle through the training set $\S{S}$, we compute the change $\delta(Y)$ of each centroid $Y$ by the distance 
\[
\delta(Y) = d(Y, Y'), 
\]
where $Y'$ is the code graph before the $t$-th cycle through the training set $\S{S}$  and $Y$ is the current code graph after the $t$-th cycle  through $\S{S}$. Based on the triangle inequality of $d$, we set the bounds according to the following rules:
\begin{align}
\label{eq:rule-for-lowerbound}
l(X, Y) &= \max\cbrace{l(X,Y') - \delta(Y), 0}\\
\label{eq:rule-for-upperbound}
u(X)    &= \min\cbrace{u(X) + \delta\args{Y_X}, \norm{\vec{x}_a - \vec{y}}},
\end{align}
Both rules guarantee that $l(X, Y)$ is always a lower bound of $d\!\args{X, Y}$ and $u(X)$ is always an upper bound of $d\!\args{X, Y_X}$. \commentout{The assertion for the lower bound follows directly from the triangle inequality of $d$. To infer that $u(X)$ is an upper bound of $d(X, Y_X)$ we assume that $u'$ is an upper bound of $d(X, Y'_X)$. Then from the triangle inequality follows that $u(X) \geq u' + \delta(Y_X)$. In addition, by definition of the graph metric $d$, we have 
\[
\norm{\vec{x}_a - \vec{y}} \geq d(X, Y_X).
\] 
Combining both inequalities yields the assertion.} Note that calculating an Euclidean distance and the computational overhead of storing aligned vector representations $\vec{x}_a$ of each training graph $X$ is computationally negligible compared to calculating the NP-hard graph distance $d$.  

Finally, we declare upper bounds as out-of-date if 
\[
\delta(Y) = d(Y, Y') > \theta
\]
where $\theta\geq 0$ is some prespecified control parameter that trades accuracy against speed. This is motivated by the following considerations. For sake of simplicity, we assume that $\theta = 0$. From 
\[
\delta(Y) = d(Y, Y') = 0
\]
follows that the current and the previous code graph are identical. This implies that an optimal alignment $(\vec{x}, \vec{y}')$ of $X \times Y'_X$  is also an optimal alignment   $(\vec{x}, \vec{y})$ of $X \times Y_X$ for all training graphs $X$ encoded by $Y_X$. According to eqn.\ (\ref{eq:rule-for-upperbound}) the upper bounds are then of the form
\[
u(X) = \norm{\vec{x} - \vec{y}} = d(X, Y_X).
\]
As a consequence, we may declare $u(X)$ as up-to-date rather then out-of-date. This makes a delayed distance evaluation of $d(X, Y_X)$ unnecessary in the next iteration. To further accelerate competitive learning GQ, we can generalize this idea for small changes 
\[
\delta(Y) = d(Y, Y') \leq \theta.
\]
The underlying assumption of this heuristic is that the more similar the previous and the recomputed code graphs are, the more likely is an optimal  alignment $(\vec{x}, \vec{y'})$ of $X \times Y'_X$ also an optimal alignment $(\vec{x}, \vec{y})$ of $X \times Y_X$. In fact, we have
\[
d(X, Y_X) \leq \norm{\vec{x} - \vec{y}} \leq \norm{\vec{x} - \vec{y'}} +\;\theta = d(X, Y'_X) + \theta
\]
showing that the upper bound $\norm{\vec{x} - \vec{y}}$ deviates at most by $\theta$ from $d(X, Y_X)$.

\introskip

Algorithm \ref{alg:ACLGQ} describes the accelerated version of competitive learning GQ. The accelerated version calls the subroutines \textsc{classify}, \textsc{update\_bounds}, and \textsc{estimate\_bounds} described in Algorithm \ref{alg:ACLGQ:classify}, \ref{alg:ACLGQ:updateBounds}, and \ref{alg:ACLGQ:estimateBounds} respectively. The subroutine \textsc{classify} determines the encoding of the current training graph by applying the principle of delayed distance evaluation. The subroutine \textsc{update\_bounds} is an auxiliary subroutine for updating the bounds and keeping book of the most recent optimal alignment of the current training graph and its encoding. Finally, \textsc{estimate\_bounds} is a subroutine that re-estimates the lower and upper bounds after each cycle through the training set.

\section{Experiments}

This section reports the results of running standard competitive learning GQ and its accelerated version.

\subsection{Data.} 
We selected four data sets described in \cite{Riesen08}. The data sets are publicly available at \cite{IAMGDB}. Each data set is divided into a training, validation, and a test set. In all four cases, we considered data from the test set only. The description of the data sets are mainly excerpts from \cite{Riesen08}. Table \ref{tab:characteristics} provides a summary of the main characteristics of the data sets.

\begin{table}[hbp]
\begin{tabular}{lcccccc}
\hline
\hline
data set & \#(graphs)& \#(classes) & avg(nodes) & max(nodes) & avg(edges) & max(edges) \\
\hline
letter & 750 & 15 & 4.7 & 8 & 3.1 & 6\\
grec & 528 & 22 & 11.5 & 24 & 11.9 & 29\\
fingerprint & 900 & 3 & 8.3 & 26 & 14.1 & 48\\
molecules & 100 & 2 & 24.6 & 40 & 25.2 & 44\\
\hline
\hline
\end{tabular}
\vspace{1ex}
\caption{Summary of main characteristics of the data sets.}
\label{tab:characteristics}
\end{table}

\begin{figure}[tbp]
\centering
\includegraphics[width=0.4\textwidth]{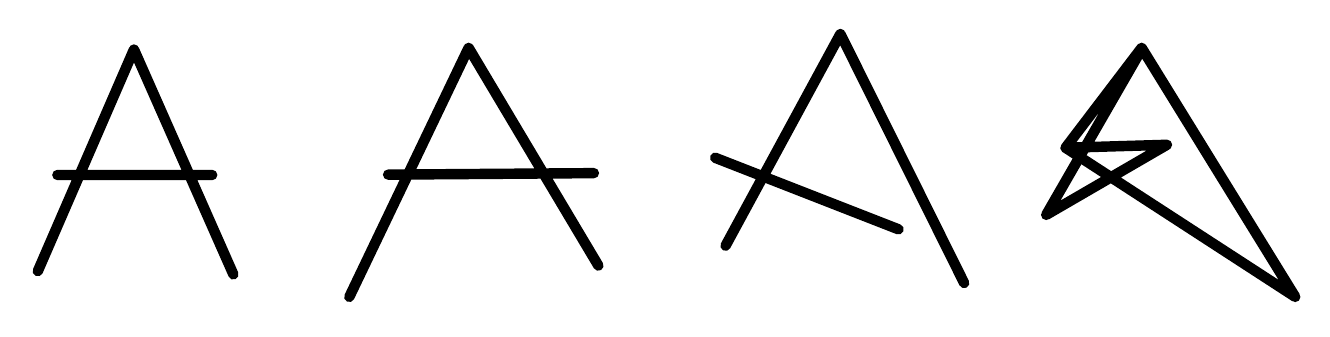}
\caption{Example of letter drawings: Prototype of letter A and distorted copies generated by imposing low, medium, and high distortion (from left to right) on prototype A.}
\label{fig:letters}
\end{figure}

\paragraph*{Letter Graphs.}
We consider all $750$ graphs from the test data set representing distorted letter drawings from the Roman alphabet that consist of straight lines only (A, E, F, H, I, K, L, M, N, T, V, W, X, Y, Z). The graphs are uniformly distributed over the $15$ classes (letters). The letter drawings are obtained by distorting prototype letters at low distortion level. Lines of a letter are represented by edges and ending points of lines by vertices. Each vertex is labeled with a two-dimensional vector giving the position of its end point relative to a reference coordinate system. Edges are labeled with weight $1$. Figure \ref{fig:letters} shows a prototype letter and distorted version at various distortion levels. 

\begin{figure}[htbp]
\centering
\includegraphics[width=0.5\textwidth]{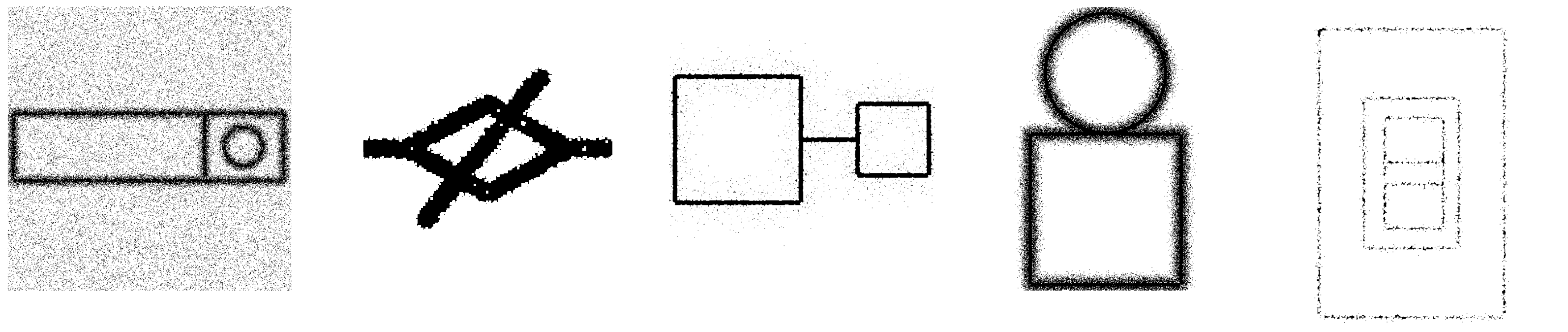}
\caption{GREC symbols: A sample image of each distortion level}
\label{fig:grec}
\end{figure}

\paragraph*{GREC Graphs.}
The GREC data set \cite{Dosch06} consists of graphs representing symbols from architectural and electronic drawings. We use all $528$ graphs from the test data set uniformly distributed over $22$ classes. The images occur at five different distortion levels. In Figure \ref{fig:grec} for each distortion level one example of a drawing is given. Depending on the distortion level, either erosion, dilation, or other morphological operations are applied. The result is thinned to obtain lines of one pixel width. Finally, graphs are extracted from the resulting denoised images by tracing the lines from end to end and detecting intersections as well as corners. Ending points, corners, intersections and circles are represented by vertices and labeled with a two-dimensional attribute giving their position. The vertices are connected by undirected edges which are labeled as line or arc. An additional attribute specifies the angle with respect to the horizontal direction or the diameter in case of arcs. 

\begin{figure}[tbp]
\centering
\includegraphics[width=0.5\textwidth]{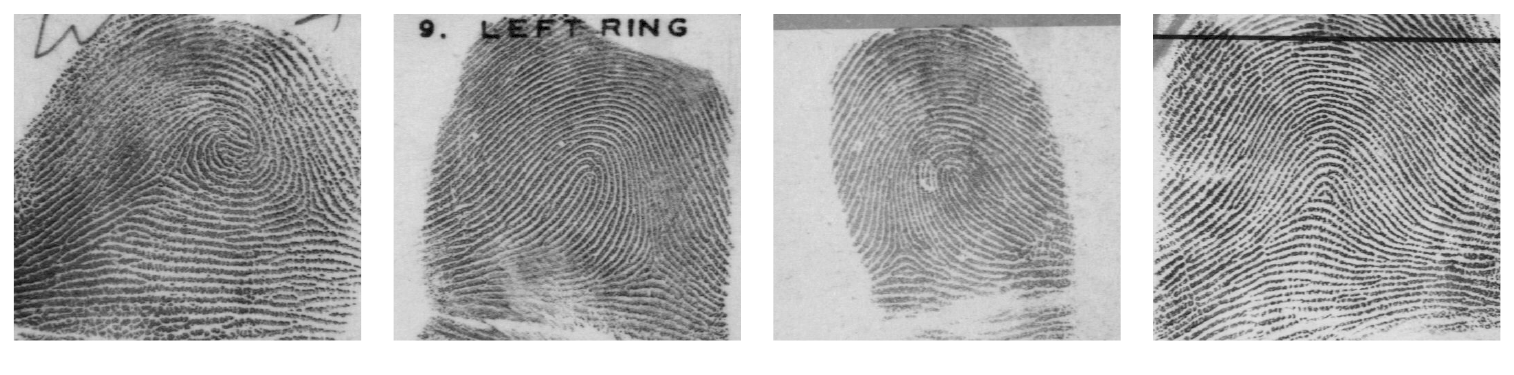}
\caption{Fingerprints: (a) Left (b) Right (c) Arch (d) Whorl. Fingerprints of class arch are not considered.}
\label{fig:fingerprints}
\end{figure}

\paragraph*{Fingerprint Graphs.}
We consider a subset of $900$ graphs from the test data set representing fingerprint images of the NIST-4 database \cite{Watson92}. The graphs are uniformly distributed over three classes \emph{left}, \emph{right}, and \emph{whorl}. A fourth class (\emph{arch}) is excluded in order to keep the data set balanced. Fingerprint images are converted into graphs by filtering the images and extracting regions that are relevant \cite{Neuhaus05}. Relevant regions are binarized and a noise removal and thinning procedure is applied. This results in a skeletonized representation of the extracted regions. Ending points and bifurcation points of the skeletonized regions are represented by vertices. Additional vertices are inserted in regular intervals between ending points and bifurcation points. Finally, undirected edges are inserted to link vertices that are directly connected through a ridge in the skeleton. Each vertex is labeled with a two-dimensional attribute giving its position. Edges are attributed with an angle denoting the orientation of the edge with respect to the horizontal direction. Figure \ref{fig:fingerprints} shows fingerprints of each class. 

\paragraph*{Molecules.}
The mutagenicity data set consists of chemical molecules from two classes (mutagen, non-mutagen). The data set was originally compiled by \cite{Kazius05} and reprocessed by \cite{Riesen08}. We consider a subset of $100$ molecules from the test data set uniformly distributed over both classes.
We describe molecules by graphs in the usual way: atoms are represented by vertices labeled with the atom type of the corresponding atom and bonds between atoms are represented by edges labeled with the valence of the corresponding bonds. We used a $1$-to-$k$ binary encoding for representing atom types and valence of bonds, respectively.

\subsection{General Experimental Setup}

In all experiments, we applied standard competitive learning GQ (std) and its accelerated version (acc) to the aforementioned data sets by using the following experimental setup:

\paragraph*{Setting of competitive learning GQ algorithms.}
 To initialize the standard and accelerated competitive learning GQ algorithm, we used a modified version of the "furthest first" heuristic \cite{Hochbaum85}. For each training set $\S{S}$, the first code graph $Y_1$ is initialized to be a graph closest to the sample mean of $\S{S}$ (see \cite{Jain08,Jain09} for details on computing the sample mean of graphs). Subsequent code graphs are initialized according to 
\[
Y_{i+1} = \arg \max_{X \in \S{S}} \min_{Y \in \S{C}_i} d(X, Y),
\]
where $\S{C}_i$ is the set of the first $i$ centroids chosen so far. We terminated both competitive learning GQ algorithms after $150$ cycles through the training set.

\paragraph*{Graph distance calculations and optimal alignment.}
For graph distance calculations and finding optimal alignments, we applied a depth first search algorithm on the letter data set and the graduated assignment \cite{Gold96a} on the grec, fingerprint, and molecule data set. The depth first search method guarantees to return optimal solutions and therefore can be applied to small graphs only. Graduated assignment returns approximate solutions. 

\paragraph*{Performance measures.}
We used the following measures to assess the performance of an algorithm on a dataset: (1) empirical distortion, (2) classification accuracy, (3) silhouette index, and (4) number of graph distance calculations.  

The \emph{empirical distortion} is here given by 
\[
\hat{D}(\S{C}) = \frac{1}{2}\sum_{i=1}^N \min_{1 \leq j \leq k} d(X_i, Y_j).
\]

The \emph{silhouette index} is a cluster validation index taking values from $[-1, 1]$. Higher values indicate a more compact and well separated cluster structure. For more details we refer to Appendix \ref{app:silhouette} and \cite{Theodoridis09}. 

Accelerated competitive learning GQ incurs computational overhead to create and update auxiliary data structures and to compute Euclidean distances. This overhead is negligible compared to the time spent on graph distance calculations. Therefore, we report number of graph distance calculations rather than clock times as a performance measure for speed. 

\subsection{Performance Comparison}
\newcommand{\rb}[1]{\raisebox{1.5ex}[-1.5ex]{#1}}
\begin{table}[tp]
\centering
\begin{footnotesize}
\begin{tabular}{l@{\qquad}l@{\qquad}c@{\quad}c@{\qquad}c@{\quad}c@{\qquad}c@{\quad}c}
\hline
\hline
data set    & measure  \\
\hline
\hline
\\[-2ex]
& & \multicolumn{2}{c@{\qquad}}{$k=15$} &  \multicolumn{2}{c@{\qquad}}{$k=30$} & \multicolumn{2}{c}{$k=45$}     \\
letter& & \multicolumn{2}{c@{\qquad}}{$N/k=50.0$} &  \multicolumn{2}{c@{\qquad}}{$N/k=25.0$} & \multicolumn{2}{c}{$N/k=16.7$}     \\
& & std & acc & std & acc & std & acc \\
\cline{2-8}
\\[-2.5ex]
&  error        & 19.2 & 19.3  & 11.1 & 11.1 & 8.0 & 8.1\\
& accuracy (in $\%$) & 72 & 72  & 90 & 90 & 94 & 94\\
& silhouette & 0.43 & 0.42 & 0.40 & 0.40 & 0.34 & 0.34\\
\cline{2-8}
\\[-2.5ex]
& matchings  $\args{\times 10^5}$ & 16.9 & 2.2 & 33.8 & 1.7 & 50.6 & 1.5\\
& speedup & 1.0 & 7.5 & 1.0 & 19.6 & 1.0 & 34.7 \\
\hline
\hline
\\[-2ex]
& & \multicolumn{2}{c@{\qquad}}{$k=22$} &  \multicolumn{2}{c@{\qquad}}{$k=33$} & \multicolumn{2}{c}{$k=44$}     \\
grec& & \multicolumn{2}{c@{\qquad}}{$N/k=24.0$} &  \multicolumn{2}{c@{\qquad}}{$N/k=16.0$} & \multicolumn{2}{c}{$N/k=12.0$}     \\
& & std & acc & std & acc & std & acc \\
\cline{2-8}
\\[-2.5ex]
&  error & 37.5 & 38.9 & 27.6 & 29.4 & 25.5 & 23.5\\
& accuracy (in $\%$) & 74 & 75 & 87 & 87 & 88 & 91\\
& silhouette & 0.40 & 0.40 & 0.42 & 0.44 & 0.45 & 0.53\\
\cline{2-8}
\\[-2.5ex]
& matchings  $\args{\times 10^5}$ & 17.4 & 3.7 & 26.1 & 3.0 & 34.8 & 2.8\\
& speedup & 1.0 & 4.7 & 1.0 & 8.7 & 1.0 & 12.2\\
\hline
\hline
\\[-2ex]
& & \multicolumn{2}{c@{\qquad}}{$k=15$} &  \multicolumn{2}{c@{\qquad}}{$k=30$} & \multicolumn{2}{c}{$k=60$}     \\
fingerprint& & \multicolumn{2}{c@{\qquad}}{$N/k=60.0$} &  \multicolumn{2}{c@{\qquad}}{$N/k=30.0$} & \multicolumn{2}{c}{$N/k=15.0$}     \\
& & std & acc & std & acc & std & acc \\
\cline{2-8}
\\[-2.5ex]
&  error  & 76.7 & 85.6 & 17.2 & 18.2 & 1.3 & 1.3\\
& accuracy (in $\%$) & 64 & 63 & 67 & 65 & 79 & 79\\
& silhouette & 0.29 & 0.29 & 0.34 & 0.33 & 0.34 & 0.36\\
\cline{2-8}
\\[-2.5ex]
& matchings  $\args{\times 10^5}$ & 20.3 & 4.5 & 40.5 & 2.0 & 81.0 & 1.2\\
& speedup & 1.0 & 4.5 & 1.0 & 19.9 & 1.0 & 66.3\\
\hline
\hline
\\[-2ex]
& & \multicolumn{2}{c@{\qquad}}{$k=2$} &  \multicolumn{2}{c@{\qquad}}{$k=6$} & \multicolumn{2}{c}{$k=10$}     \\
molecules& & \multicolumn{2}{c@{\qquad}}{$N/k=50.0$} &  \multicolumn{2}{c@{\qquad}}{$N/k=16.7$} & \multicolumn{2}{c}{$N/k=10.0$}     \\
& & std & acc & std & acc & std & acc \\
\cline{2-8}
\\[-2.5ex]
&  error & 61.9 & 62.1 & 55.7 & 54.8 & 52.6 & 52.9 \\
& accuracy (in $\%$) & 64 & 62 & 66 & 66 & 66 & 68\\
& silhouette & 0.06 & 0.06 & 0.03 & 0.03 & 0.01 & 0.00\\
\cline{2-8}
\\[-2.5ex]
& matchings  $\args{\times 10^4}$ & 6.0 & 5.7 & 18.0 & 11.2 & 30.0 & 12.3\\
& speedup & 1.0 & 1.1 & 1.0 & 1.6 & 1.0 & 2.4\\
\hline
\hline
\end{tabular}
\end{footnotesize}
\vspace{2ex}
\caption{Results of standard competitive learning GQ (std) and accelerated competitive learning GQ (acc) on four data sets. Rows labeled \emph{matchings} give the number of graph distance calculations, and rows labeled \emph{speedup} show how many times an algorithm is faster than standard competitive learning GQ for graphs.}
\label{tab:results}
\end{table}

We applied both competitive learning GQ algorithms to all four data sets in order to assess and compare their performance. The control parameter $\theta$ of the accelerated version is set to zero. For each data set $10$ runs of each algorithm were performed and the best result with respect to the error value  (empirical distortion) is selected. For taking the average, $10$ runs are too less in order to provide statistically significant results. For conducting a higher number of runs in order to obtain statistical significant results, computational resources  were not sufficient.

Table \ref{tab:results} summarizes the results. The first observation to be made is that the solution quality of standard and accelerated competitive learning GQ is comparable with respect to error, classification accuracy, and silhouette index. Deviations are due to the non-uniqueness of the solutions and the approximation errors of the graduated assignment algorithm. The second observation to be made from Table \ref{tab:results} is that the accelerated version outperforms standard competitive learning GQ with respect to computation time on all data set. 

The results show that the speedup factor increases with increasing number $k$ of codebook graphs but obviously is independent of the average cluster size $N/k$. Contrasting the silhouette index and the dimensionality of the data to the speedup factor gained by accelerated competitive learning GQ, we make the following observation: First, the silhouette index for the letter, grec, and fingerprint data set are roughly comparable and indicate a cluster structure in the data, whereas the silhouette index for the molecule data set indicates almost no compact and homogeneous cluster structure. Second, the dimensionality of the vector representations is largest for molecule graphs, moderate for grec graphs, and relatively low for letter and fingerprint graphs. Thus, the speedup factor of accelerated competitive learning GQ apparently decreases with increasing dimensionality and decreasing cluster structure. This behavior is in line with Elkan's k-means for vectors \cite{Elkan03} and Elkan's k-means for graphs \cite{Jain09c} as well as with findings in high-dimensional vector spaces. According to \cite{Moore00}, there will be little or no acceleration in high dimensions if there is no underlying structure in the data. This view is also supported by theoretical results from computational geometry \cite{Indyk99}.

\begin{figure}[tbp]
\centering
\includegraphics[width=0.6\textwidth]{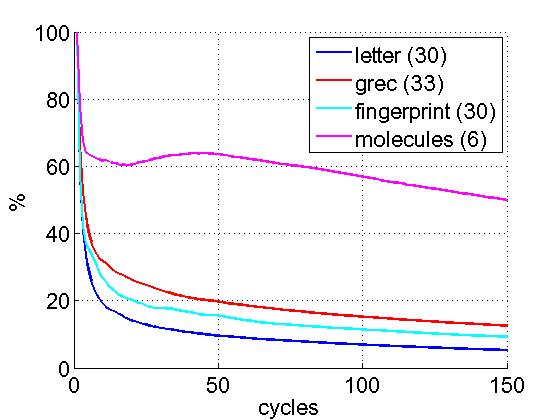}
\caption{Percentage of graph distances calculated by accelerated competitive learning GQ at each cycle through the training set. The standard version of competitive learning GQ requires $100 \%$ (corresponding to $kN$) graph distance calculations at each cycle. The number $k$ for each training set is shown in parentheses next to the identifiers of the data sets in the legend box.}
\label{fig:clgq}
\end{figure}

 Figure \ref{fig:clgq} shows how the number of graph distance calculations of accelerated competitive learning GQ decreases with increasing number of cycles through the training set. The standard  version of competitive learning GQ requires $kN$ graph distance calculations at each cycle corresponding to $100\%$. For the letter, grec, and fingerprint data set, the plot shows that after a few cycles more than $80\%$ out of $kN$ graph distance calculations can be avoided and replaced by calculations of simple Euclidean distances. Accelerated competitive learning GQ operates in both spaces, the Euclidean graph space and its underlying Euclidean representation space. The graph space is used for choosing vector representations optimally aligned to their code graphs. Given optimal aligned vector representations, competitive learning GQ reduces to competitive learning VQ in the representation space. Switching back to the graph space serves as correcting the optimal alignments between the vector representations of the input and their closest code graphs. As the plot shows, the corrective function of the graph space is less required with increasing number of cycles through the training set.

\section{Conclusion}

Accelerated competitive learning GQ avoids graph distance calculations by exploiting the metric properties of the graph distance and by lifting input graphs from the training set to vector representations that are optimally aligned to their encodings. In doing so, accelerated competitive learning GQ switches between two spaces, the graph space and its underlying Euclidean representation space. The competitive learning in the graph space serves as a corrective of the fast but erroneous counterpart in the representation space. With increasing time competitive learning for GQ gradually turns into competitive learning for VQ without loss of solution accuracy. In particular, during the long final convergence phase most graph distance calculations can be avoided and replaced by Euclidean distances, provided the data possesses a cluster structure. We conclude that accelerated competitive learning GQ is a first step to avoid intractable graph distance calculations without loss of structural information. We believe that the proposed acceleration can also be applied to GQ using more general graph edit distance metrics as underlying distortion measures. As a next step, future work is concerned with an empirical investigation of the lifting threshold $\theta$ to trade solution quality against speed.

\begin{appendix}

\section{The Silhouette Index}\label{app:silhouette}
Suppose that $\S{S} = \cbrace{X_1, \ldots, X_m}$ is a sample of $m$ patterns. 
Let $\S{C} = \cbrace{\S{C}_1, \ldots, \S{C}_k}$ be a partition of $\S{S}$ consisting of $k$ disjoint clusters with 
\[
\S{S} = \bigcup_{i=1}^{k}\S{C}_i.
\] 
We assume that $D$ is the underlying distance function defined on $\S{S}$. The distance between two subsets $\S{U}, \S{U}' \subseteq \S{S}$ is defined by 
\[
D\args{\S{U}, \S{U}'} = \min\cbrace{D\args{X, X'} \,:\, X \in \S{U}, X'\in \S{U}'}.
\]
If $\S{U} = \cbrace{X}$ consists of a singleton, we simply write $D\args{X, \S{U}'}$ instead of $D\args{\cbrace{X}, \S{U}'}$. 

Let
\[
D_{\text{avg}}\args{X, \S{U}} 
\]
denote the average distance between pattern $X \in \S{S}$ and subset $\S{U} \subseteq \S{S}$. Suppose that pattern $X_i \in \S{S}$ is a member of cluster $\S{C}_{m(i)} \in \S{C}$. By $\S{C}'_{m(i)}$ we denote the set $\S{C}_{m(i)} \setminus\cbrace{X_i}$. For each pattern $X_i \in \S{S}$ let
\[
a_i = D_{\text{avg}}\args{X, \S{C}'_{m(i)}}
\] 
be the average distance between pattern $X_i$ and subset $\S{C}'_{m(i)}$. By 
\[
b_i = \min_{j \neq m(i)} D_{\text{avg}}\args{X_i, \S{C}_j}
\]
we denote the minimum average distance between pattern $X_i$ and all clusters from $\S{C}$ not containing $X_i$. The \emph{silhouette width} of $X_i$ is defined as
\[
s_i = \frac{b_i - a_i}{\max\args{b_i, a_i}}. 
\] 
The \emph{silhouette of cluster} $\S{C}_j \in \S{C}$ is given by
\[
S_j = \frac{1}{\abs{\S{C}_j}} \sum_{i:X_i\in\S{C}_j} s_i.
\]
The \emph{silhouette index} is then defined as the average of all cluster  silhouettes 
\[
\mathfrak{S} = \frac{1}{k} \sum_{j=1}^{k} S_j.
\]
\end{appendix}

\bibliographystyle{splncs}

\end{document}